\documentclass[10pt,twocolumn,letterpaper]{article}

\usepackage[pagenumbers]{wacv} 
\usepackage{amsmath}
\usepackage{pifont}
\usepackage{fancybox}
\usepackage{threeparttable}
\usepackage{pifont}
\usepackage{multirow}
\usepackage{graphicx}

\definecolor{wacvblue}{rgb}{0.21,0.49,0.74}
\usepackage[pagebackref,breaklinks,colorlinks,allcolors=wacvblue]{hyperref}

\newcommand{\squared}[1]{\raisebox{0.3ex}{\fbox{\raisebox{-0.4ex}[0.6ex][0pt]{\makebox[0.6ex]{#1}}}}}

\newcommand{\circled}[1]{%
    \raisebox{0em}{\textcircled{\raisebox{-0.05em}{\bf \footnotesize $#1$}}}%
}

\def\invcircledast#1{%
  \mathbin{\vphantom{\circledast}\text{%
    \ooalign{\smash{\blackcircle}\cr
             \hidewidth\smash{\textcolor{white}{\bf \footnotesize $#1$}}\hidewidth\cr
            }%
  }}%
}
\newcommand{\blackcircle}{\raisebox{-.6ex}{\hspace{-0.1ex}\scalebox{2.30}{\raisebox{-0.05ex}{$\bullet$}}}}

\newcommand{\mname}{TellTrack }
\newcommand{\Mname}{TellTrack}

\def\wacvPaperID{} 
\def\confName{WACV}
\def\confYear{2026}

\title{\underline{Tell} Me What to \underline{Track}: Infusing Robust Language Guidance for Enhanced Referring Multi-Object Tracking}

\author{Wenjun Huang$^{1}$,
Yang Ni$^{1}$, Hanning Chen$^{1}$, Yirui He$^{1}$, Ian Bryant$^{1}$, Yezi Liu$^{1}$, Mohsen Imani$^{1}$\\
$^{1}$ University of California, Irvine, CA, USA\\
{\tt\small \{m.imani\}@uci.edu}
\thanks{
This work was supported in part by the DARPA Young Faculty Award, the National Science Foundation (NSF) under Grants \#2127780, \#2319198, \#2321840, \#2312517, and \#2235472, the Semiconductor Research Corporation (SRC), the Office of Naval Research through the Young Investigator Program Award, and Grants \#N00014-21-1-2225 and \#N00014-22-1-2067, and Army Research Office Grant \#W911NF2410360. Additionally, support was provided by the Air Force Office of Scientific Research under Award \#FA9550-22-1-0253, along with generous gifts from Xilinx and Cisco.
}
}

\begin{document}
\maketitle

\begin{abstract}
Referring multi-object tracking (RMOT) is an emerging autonomous driving task that aims to localize an arbitrary number of targets based on a language expression and continuously track them in a video. This intricate task involves reasoning on multi-modal data and precise target localization with temporal association. However, prior studies overlook the imbalanced data distribution between newborn targets and existing targets due to the nature of the task. In addition, they only indirectly fuse multi-modal features, struggling to deliver clear guidance on newborn target detection. To solve the above issues, we propose \textbf{\Mname}. \mname conducts a collaborative matching strategy to alleviate the impact of the imbalance, boosting the ability to detect newborn targets while maintaining tracking performance. In the encoder, we integrate and enhance the cross-modal and multi-scale fusion, overcoming the bottlenecks where limited multi-modal information is shared and interacted between feature maps. In the decoder, we also develop a referring-infused adaptation that provides explicit referring guidance through the query tokens. The experiments on the according autonomous driving datasets showcase our superior performance (+3.22\%) compared to prior works, demonstrating the effectiveness of our designs.
\vspace{-2em}
\end{abstract}

\section{Introduction}
\begin{figure}[t]
  \centering
   \includegraphics[width=\columnwidth]{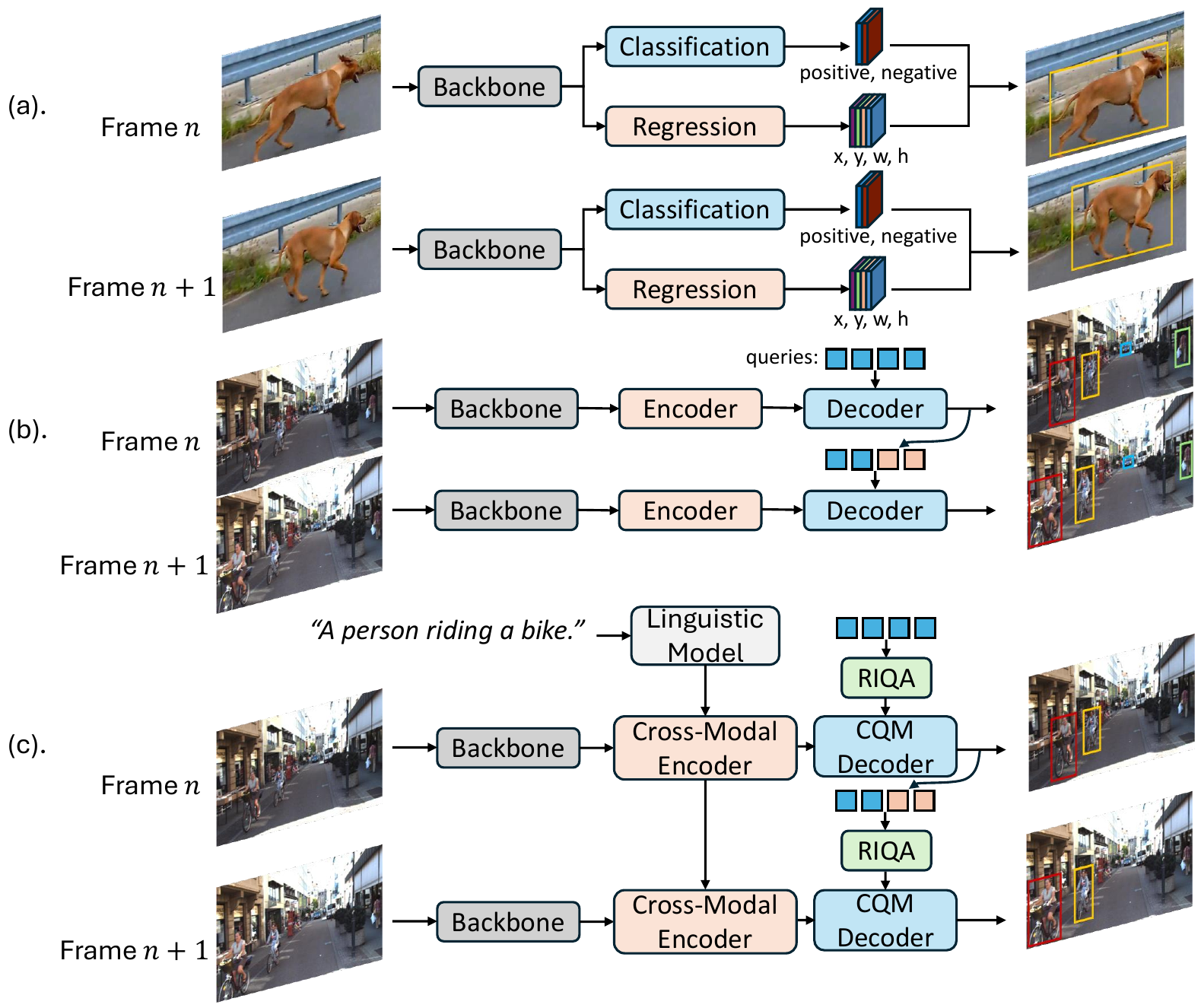}
   \caption{
   Tracking pipelines for SOT, MOT, and our \mname for RMOT. (a). SOT focuses on tracking one object and maintaining its location with a classification head and a regression head. (b). MOT tracks multiple objects simultaneously and manages multiple identities. (c).\mname considers a language prompt (``A person riding a bike'' in this case) and only tracks the targets meeting the description.
   \mname consists of a novel referring-infused query adaptation (RIQA) module, a collaborative query matching (CQM) decoder, and a cross-modal encoder (CME). 
   }
   \vspace{-2.2em}
   \label{fig: motivation}
\end{figure}

Over the past decades, multi-object tracking (MOT) has played an important role in autonomous driving \cite{luo2021exploring}, 
where understanding and monitoring the movement of multiple entities over time is critical.
MOT can be defined as the process of following the trajectories of a set of objects through different frames while keeping their identities distinct.
Compared with single object tracking (SOT) \cite{bolme2010visual, bertinetto2016fully, fan2019lasot}, MOT not only detects and associates more objects but also manages unique identities for each object and handles frequent occlusions.

Traditional MOT methods often lack the nuanced understanding required to follow specific targets when guided by natural language descriptions, a challenge that becomes particularly evident when users wish to focus on objects of interest described semantically. Meanwhile, referring understanding \cite{yu2016modeling, khoreva2019video, deruyttere2019talk2car, wu2023referring} that integrates natural language processing into scene perception has raised great attention, with the advancement of vision-language models (VLMs).
It aims to localize targets of interest in images or videos under the instruction of human language.
In this paper, we focus on an emerging task named referring multi-object tracking (RMOT), which enhances conventional MOT and takes into account language understanding.
RMOT improves MOT's ability to meet human intentions, significantly broadens the applicability, and boosts functional efficiency. Instead of tracking all visible objects in the scene, RMOT aims at tracking only the referent targets.
For example, if we input ``A person riding a bike'' as the text prompt, the tracker should only track the ones meeting the description while ignoring other objects such as ``cars'' and ``a person on foot'', which are also tracked in MOT.
\cref{fig: motivation} illustrates the typical tracking pipelines for SOT, MOT, and our proposed pipeline for RMOT. 

Nevertheless, current transformer-based models are faced with several challenges that lead to a sub-optimal performance in RMOT~\cite{zhang2024bootstrapping, wu2023referring}.
First of all, based on the transformer architecture, they train a joint decoder for newborn target detection and existing target tracking.
However, the imbalanced distribution of newborn targets and existing targets in the dataset impairs the training of newborn target detection.
While the query tokens for existing targets, referred to as ``track queries'' are activated and trained during the whole lifespan of the targets, the queries for newborn targets, referred to as ``detection queries'', are only activated once when the targets first appear in the video.
This insufficient training in newborn target detection leads to poor performance when dealing with uncommon targets.

In terms of language guidance, current designs fuse the text embedding with image features right after the vision backbone, providing a mixed-modal feature map to later stages. However, the fused feature is not a direct input to the most critical decoder and contains no explicit semantic information, leading to relatively weak and indirect language guidance that cannot be effectively reasoned in the decoder.
Recent work, such as Segment Anything (SAM) \cite{kirillov2023segment, ravi2024sam}, adopts a different paradigm to fuse images and prompts.
Specifically, SAM concatenates the prompt embedding and query tokens as the input to the decoder, providing strong and direct guidance. 
However, models like SAM cannot be naively applied in RMOT.
Despite its strong zero-shot segmentation performance, SAM requires explicit point or box prompts to focus on a specific instance and does not support arbitrary language guidance or newborn target detection, not fitting the need of the RMOT task.

Observing the aforementioned challenges, we propose a new RMOT algorithm, named ``\underline{Tell} Me What to \underline{Track}'' (\Mname).
On one hand, \mname proposes a strategy that relaxes the matching criteria and thereby increases the activation frequency of the detection queries.
During training, the existing targets are not solely matched with track queries but can also be matched with detection queries.
On the other hand, \mname adopts a query adaptor that directly fuses the text prompt with the queries, providing strong guidance and enhancing the model's reasoning capability.
Before the decoder, \mname also develops a unified encoder that generates a well-rounded fusion of both modalities and effectively incorporates interaction among multi-scale feature maps and text inputs.
The key contributions of this work are outlined as follows:
\begin{itemize}
    \item \mname introduce an effective training strategy to boost the model's detection performance by jointly training the detection queries and track queries, which alleviates the impact caused by the imbalanced distribution of targets.
    \item Prior works leverage a limited architecture to enable referring in MOT tasks, leading to weak multi-modal fusion and difficulties in understanding nuanced user intention. In contrast, \mname provides stronger and more direct guidance to ensure more accurate tracking. 
    \item \mname redesign the decoder's architecture to better integrate the text prompt into the decoder queries; outside of the decoder, we develop a new cross-modal encoder that boosts the information exchange between the multi-modal and multi-scale features. 
    \item Extensive experiments confirm the superior performance of \Mname, leading to a +3.22\% improvement.
\end{itemize}
\vspace{-1em}
\section{Preliminaries}
\label{subsec: arch}
\vspace{-0.5em}
Taking the video stream and a language query as inputs, the goal of an RMOT model is to output the track boxes of the corresponding query. 
A plain transformer-based RMOT model \cite{wu2023referring}, building on top of Deformable DETR \cite{zhu2020deformable} and MOTR \cite{zeng2022motr}, consists of four key components: feature extractor, encoder, decoder, and temporal reasoning module. 

The feature extractor first produces visual and linguistic features for the raw video and text. 
Given a $N$-frame video, an image backbone extracts the frame-wise feature pyramid $\textbf{I}_n^l \in \mathbb{R}^{C_l\times H_l \times W_l}$, where $n$ represents the frame index,  and $C_l, H_l, W_l$ represents the channel depth, height, width of the $l^{\text{th}}$ level feature map, respectively. 
At the same time, a linguistic model embeds the text description $T$ into a set of word embeddings $\textbf{S}_w \in \mathbb{R}^{L\times D}$, where $D$ is the embedding dimension, and $L$ is the number of embedded tokens.
Then, an encoder fuses the features of two modalities per frame and gets a stack of vision-language fused embeddings $E_n=\{E_n^1, \cdots, E_n^l\}$. For each level of feature maps:
\begin{align}
    E_n^l = \text{Attn}(
    Q=\text{PE}^I(\textbf{I}_n^l), 
    K=\text{PE}^S(\textbf{S}_w),
    V=\textbf{S}_w) \label{eq: F_fuse}
\end{align},
where $\text{PE}^I(\cdot), \text{PE}^S(\cdot)$ integrate the positional embeddings into image features and text features, respectively. ``Attn'' refers to a few attention blocks.
\begin{figure*}[t]
  \centering
   \includegraphics[width=0.9\linewidth]{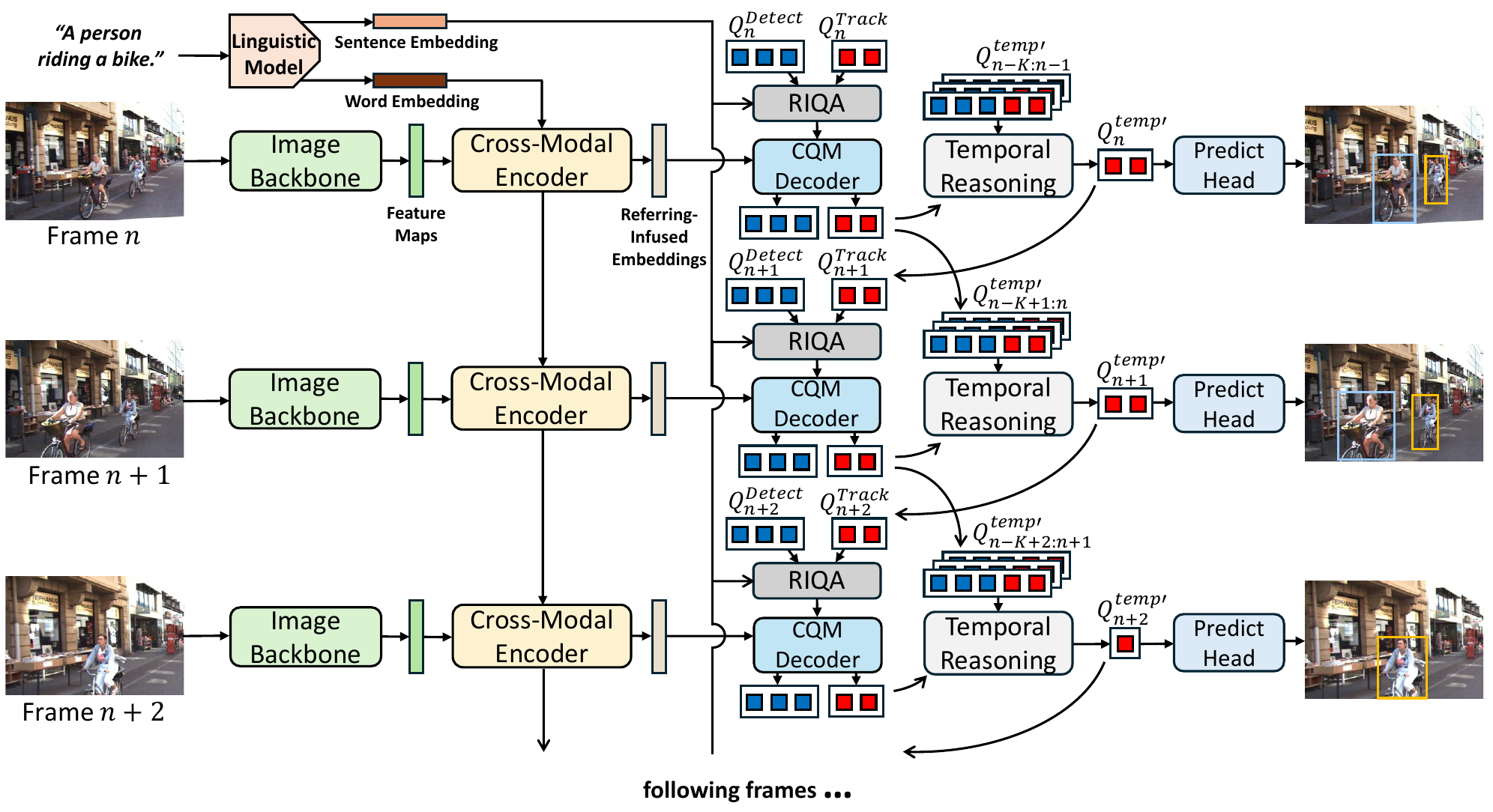}

   \caption{An overview of \Mname. The transformer-based framework with a memory bank accepts a video frame, a language expression, and a set of learnable queries as input. With the temporal information from past frames in the memory bank, it outputs embeddings corresponding to the tracked targets.
   RIQA fuses the language information with queries and CQM jointly optimizes the newborn target detection and existing target tracking.}
   \label{fig: overview}
   \vspace{-1.5em}
\end{figure*}
A deformable attention encoder \cite{zhu2020deformable} is then adopted to further refine the embeddings.
\begin{equation}
    E_n^\prime = \text{MSDeformAttn}(E_n, p_q, E_n)
    \label{eq: deform_enc}
\end{equation}
, where ``MSDeformAttn'' follows the notation in \cite{zhu2020deformable}, and $p_q$ denotes the reference points for deformable attention.

Next, a decoder is used to process a set of learnable queries $Q_n = \{Q^{Det}_n, Q^{Tra}_n\}$ for object detection and tracking.
$Q_n$ are categorized into two types: 
$Q^{Det}$ detects potential newborn targets in the current frame, and $Q^{Tra}$ represents the tracked targets from the previous frames that aim to locate the same target in the current frame.
After self-attending to other queries and cross-attending to the multi-modal embeddings $E_n^\prime$, the updated queries $Q_n^\prime = \{Q^{Det\prime}_n, Q^{Tra\prime}_n\}$ are fed into a referent head (RH) for newborn target detection and existing target tracking.
The RH consists of three branches: class, box, and referring. 
The class branch uses a linear projection to output a binary probability ($\{\hat{c}^{Det}_n$, $\hat{c}^{Tra}_n\}$), indicating whether the resulting embedding represents a real object. 
The box branch is a 3-layer feed-forward network (FFN) with ReLU activations, except for the final layer, and predicts the bounding box location ($\{\hat{b}^{Det}_n$, $\hat{b}^{Tra}_n\}$) for all visible objects.
The referring branch is another linear projection that outputs referent scores ($\{\hat{r}^{Det}_n$, $\hat{r}^{Tra}_n\}$) as binary values, reflecting the likelihood that the object matches the given expression.

After decoding the queries from single-frame features, the temporal reasoning module integrates the information from past frames per query, in order to refine each query and its predicted box. 
It can be formulated as follows:
\vspace{-.5em}
\begin{align}
    q^{temp}_n = &\text{Attn}(
     Q=\text{PE}^T(q_n^\prime),\notag \\ 
    &K=\text{PE}^T(q_{n-K:n-1}^{temp\prime}), V=q_{n-K:n-1}^{temp\prime})
\end{align}
, where $q_n^\prime$ is individual query in $Q_n^\prime$, $q_{n-K:n-1}^{temp\prime}$ are the temporal refined queries of $q_n^\prime$ in previous $K$ frames; $\text{PE}^T$ represents temporal positional encoding. All temporal refined queries form $Q_n^{temp}=\{q_{n, 1}^{temp},\cdots,q_{n, m}^{temp}\}$.
\begin{align}
    Q^{temp\prime}_n = &\text{Attn}(
    Q=\text{PE}^Q(Q^{temp}_n),\notag \\ 
    &K=\text{PE}^Q(Q^{temp}_n), V=Q^{temp}_n)
\end{align},
where $\text{PE}^Q$ represents query positional embedding.

Given the temporally refined queries, we calculate the offsets to refine the boxes and get the final predictions $\{\hat{b}_n^{Det\prime}, \hat{b}_n^{Tra\prime}\}$.
\begin{equation}
    \Delta \hat{b}_n^{Det}, \Delta \hat{b}_n^{Tra} = \text{FFN}(Q_n^{temp\prime})
\end{equation}
\begin{equation}
    \hat{b}_n^{Det\prime}, \hat{b}_n^{Tra\prime} = \Delta \hat{b}_n^{Det} + \hat{b}_n^{Det}, \Delta \hat{b}_n^{Tra} + \hat{b}_n^{Tra}
\end{equation}
\vspace{-1.5em}

\section{Method}\label{sec: method}
In this section, we elaborate on each newly proposed component in \Mname, which can readily enhance the performance with a negligible computation overhead: \textbf{(1). Collaborative Query Matching (CQM)}. The imbalance of newborn targets and existing targets impairs the overall model performance.
We use CQM to facilitate newborn object detection while maintaining the performance of existing target tracking.
\textbf{(2). Referring-Infused Query Adaptation (RIQA)}. 
In addition to the indirect fusion of the language description and image in the encoder, we inject a direct information change between the reference and the queries in the decoder, which explicitly guides the queries to detect desired targets.
\textbf{(3). Cross-Modal Encoder (CME)}.
The encoder of the previous work suffers from the limited perceptive field of the image features.
We develop a new CME to boost the multi-modal fusion by facilitating the information exchange between the multi-scale feature maps and the text.  
The overview of our method is illustrated in \cref{fig: overview}.

\subsection{Collaborative Query Matching} \label{subsec: cqm}

\begin{figure}[t]
  \centering
   \includegraphics[width=\columnwidth]{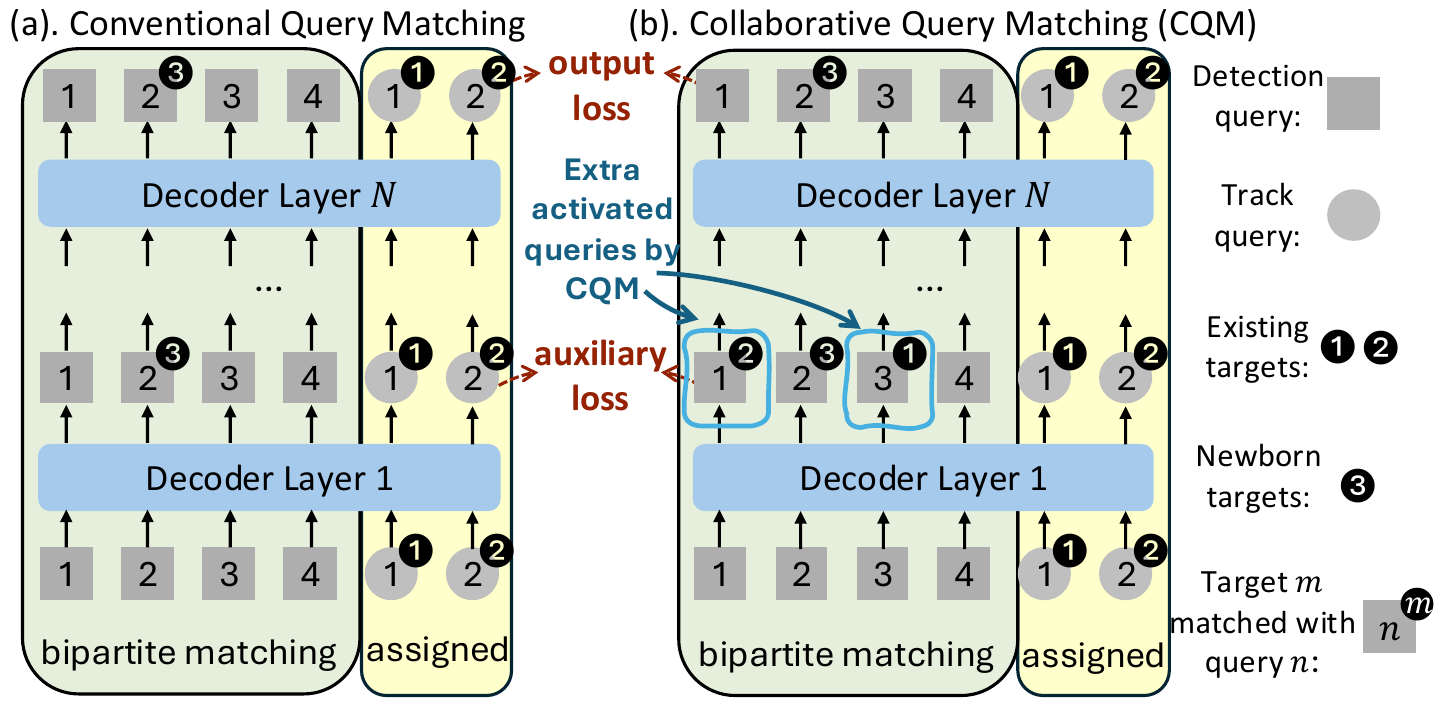}
    \vspace{-2em}
   \caption{Comparison of conventional query matching and CQM. 
   In both matching, track queries match the pre-assigned existing targets at each decoder layer. Conventional matching performs a bipartite matching between detection queries and newborn targets at each layer. In contrast, CQM matches detection queries with both existing and newborn targets in the intermediate decoder layers, except in the final layer.
   Both methods compute the output loss similarly. However, in CQM, existing targets are also included in the \textit{detection loss} during auxiliary loss calculation (See \cref{subsec: loss}).
   }
   \vspace{-1.5em}
   \label{fig: cola}
\end{figure}

Traditional transformer-based MOT algorithms adopt one-to-one bipartite matching for detection queries in all decoder layers. 
However, the algorithms underestimate the imbalanced activations between detection queries and track queries caused by the natural difference in the number of newborn targets and existing targets in the dataset. That is, any object that newly appears will become existing targets in later frames, making newborn targets relatively sparse in the dataset. 
Once the target is detected, it is assigned to a track query in the following frames, therefore, each target only activates the detection query once but activates the track query multiple times in the subsequent frames.
The insufficient training of detection queries significantly impairs the model's performance.
To resolve the interference in newborn detection caused by track queries, we propose CQM. 

In current auxiliary training, as the example depicted in \cref{fig: cola}, the intermediate outputs of each decoder layer are also treated as final outputs and contribute to the loss. 
In each layer, track queries (\circled{1} and \circled{2}) are trained to localize the same pre-matched targets ($\invcircledast{1}$ and $\invcircledast{2}$), and detection queries do a one-to-one bipartite matching with those newborn targets ($\invcircledast{3}$).
Therefore, in this example, only one detection query ( \squared{2} ) is effectively activated and trained.

Instead of urging a one-to-one matching, CQM allows the detection queries to match the existing targets in the intermediate layers, as shown in \cref{fig: cola}.
Except for the final output, the existing targets can be discovered by another detection query besides the assigned track query.
In the example, $\invcircledast{1}$ and $\invcircledast{2}$ are not only matched with \circled{1} and \circled{2}, respectively, but also are matched with detection queries \squared{3} and \squared{1}.
Therefore, in addition to \squared{2} is trained by matching with $\invcircledast{3}$,  \squared{1} and \squared{3} are also activited.
Compared with traditional auxiliary training, CQM significantly boosts the training frequency of detection queries and, as a consequence, improves the model performance.

\subsection{Referring-Infused Query Adaptation} \label{subsec: riqa}
Recent works \cite{wu2023referring, zhang2024bootstrapping} focus on fusing the text prompt with image features in the early stages of the system, which lacks direct guidance for object detection in the later decoder stage.
To tackle this, we encode the user's semantic intention directly into the queries to provide explicit guidance.
The organization of our queries, which follows previous work \cite{zhu2020deformable, liu2022dab, wang2022anchor, meng2021conditional}, consists of two parts: \textit{position part}, and \textit{content part}.
The position part presents the spatial prior, and the content part represents the semantic prior of the query.
Each query has a dimension of $\mathbb{R}^{1\times 2D}$.
The first half of each query is the position part and the second half of each is the content part.
Inspired by the decoder architectures of Deformable-DETR \cite{zhu2020deformable} and SAM \cite{kirillov2023segment}, we propose two different types of RIQAs, i.e., pre-decoder adaptation and in-decoder adaptation, that inject sentence embedding into the \textit{content part} of each query.
Both the pre-decoder adaptation and the in-decoder adaptation first generate a sentence embedding $\textbf{S}_s \in \mathbb{R}^{1\times D}$ of the text prompt $T$ via a frozen sentence encoder and a trainable FFN.
\begin{equation}
    \textbf{S}_s = \text{FFN}(\text{SentenceEncoder}(T))
\end{equation}
For infusing referring text, we especially choose sentence embedding over individual word embeddings to provide more general, flexible, and meanwhile less restricted guidance in the query.

\begin{figure}[t]
  \centering
    \includegraphics[width=0.9\columnwidth]{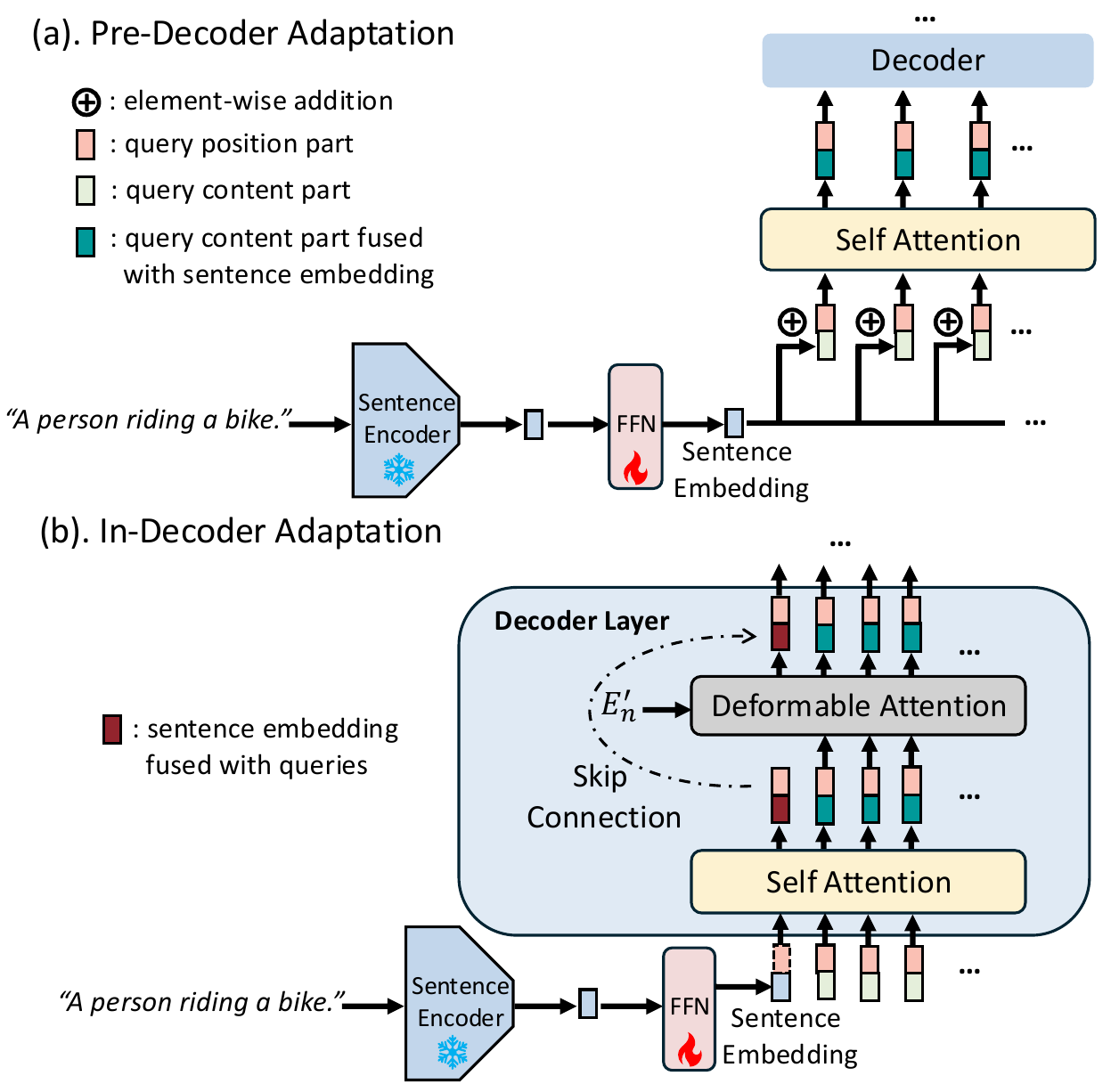}
    \vspace{-.5em}
   \caption{Overview of Referring-Infused Query Adaptation: (a) Pre-Decoder Adaptation and (b) In-Decoder Adaptation. Pre-decoder adaptation integrates sentence embeddings with queries before entering the decoder with an extra self-attention layer. In-decoder adaptation integrates them within the decoder by leveraging the existing self-attention stage. Notice that the sentence embeddings in (b) only participate in self-attention, bypassing the deformable attention between queries and cross-modal features.}
   \vspace{-1.7em}
   \label{fig: tqf}
\end{figure}

\subsubsection{Pre-Decoder Adaptation}
The overview of pre-decoder adaptation is depicted in \cref{fig: tqf} (a).
It first fuses linguistic intention with the queries, then feeds the referring-infused queries into the decoder, the same as in previous work.
Formally, element-wise $\textbf{S}_s$ is added to the content part of each query from the last frame $Q_{n-1}[content]$.
\begin{equation}
    Q_{n-1}[content] = Q_{n-1}[content] \oplus \textbf{S}_s
\end{equation}
, where $\oplus$ represents element-wise addition.
The referring-infused queries for the current frame $Q_n$ are obtained through a self-attention layer outside the decoder.
\begin{align}
    Q_n = \text{Attn}(
    & Q=\text{PE}^T(Q_{n-1}),\notag \\ 
    & K=\text{PE}^T(Q_{n-1}),\notag \\
    & V=Q_{n-1})
\end{align}

\subsubsection{In-Decoder Adaptation}
The in-decoder adaptation uses a different way to fuse linguistic intention with queries, as depicted in \cref{fig: tqf} (b).
The sentence embedding $\textbf{S}_s$ first passes through a FFN and then is concatenated with a trainable position part $q_p \in \mathbb{R}^{1\times D}$ to form a referring query $S_s^\prime\in \mathbb{R}^{1\times 2D}$.
Next, we concatenate this extra query with the original detection and track queries to form a new set of queries $Q_{n_{adapt}}=\{S_s^\prime, Q_n\}$ for in-decoder adaptation, inspired by \cite{jia2022visual}.
Taking the concatenated queries $Q_{n_{adapt}^{j-1}}$, each decoder layer $j$ computes the outputs as follows.
We first fuse the information across the queries via self-attention.
\begin{align}
    Q_{n_{adapt}^{j\prime}} = \text{Attn}(&Q=PE^Q(Q_{n_{adapt}^{j-1}}),\notag \\ &K=PE^Q(Q_{n_{adapt}^{j-1}}),\notag \\ &V=Q_{n_{adapt}^{j-1}})
\end{align}
It can be decoupled into two parts $Q_{n_{adapt}^{j\prime}}=\{\textbf{S}_{s^{j\prime}}^\prime, Q_{n^{j\prime}}\}$.
Since the deformable attention has a constraint on the number between the reference points and the queries \cite{zhu2020deformable}, we only do deformable attention between the language fused embeddings $E_n^\prime$ and the non-linguistic queries, i.e., $Q_{n^{j\prime}}$.
\begin{equation}
    Q_{n_{deform}^{j\prime}} = \text{MSDeformAttn}(Q_{n^{j\prime}}, p_q, E_n^\prime)
\end{equation}
The output queries of the decoder layer $j$ are obtained by catenating $\textbf{S}_s^\prime$ with $Q_{n_{deform}^{j\prime}}$ and forwarding to an FFN with a residual connection.
\begin{equation}
    Q_{n_{adapt}^{j}} = \text{FFN}(\text{Concate}(\textbf{S}_s^\prime, Q_{n_{deform}^{j\prime}}) + Q_{n_{adapt}^{j\prime}})
\end{equation}

\subsection{Cross-Modal Encoder}
\begin{figure}[t]
  \centering
    \includegraphics[width=\columnwidth]{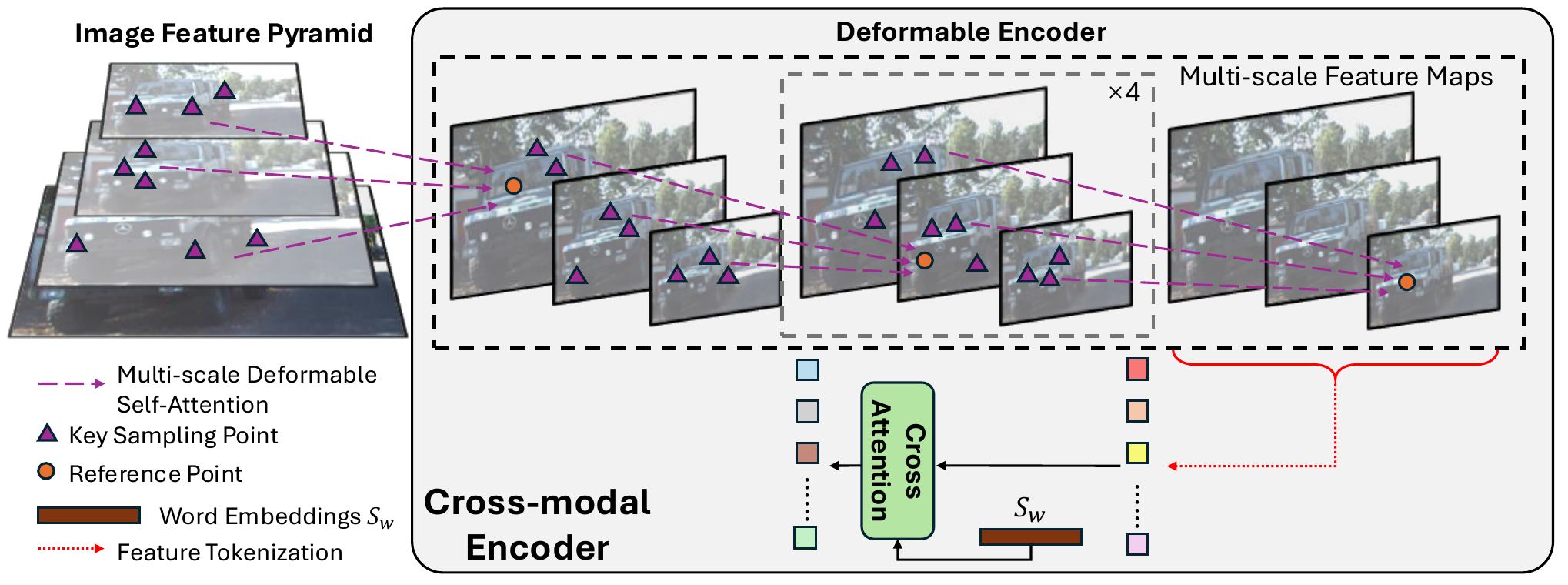}
   \caption{Overview of Cross-modal Encoder (CME). The image feature pyramid is processed by deformable self-attention layers to capture spatial dependencies with key sampling and reference points. The CME then aligns visual and textual information through cross-attention mechanisms, enabling enhanced feature representation learning from word embeddings $\textbf{S}_w$. }
   \vspace{-1em}
   \label{fig: fuse}
\end{figure}

As introduced in \cref{subsec: arch}, the previous encoder is constructed by a multi-modal fuser (\cref{eq: F_fuse}) followed by a deformable encoder (\cref{eq: deform_enc}).
However, this is suboptimal because the image feature pyramid at this stage is unstructured, noisy, and lacks hierarchical organization \cite{dosovitskiy2020image, carion2020end}. 
Since convolutional or transformer-based feature extractors generate dense feature pyramid with redundant and low-level details, and on the other hand, textual representations $(\textbf{S}_w)$ tend to be abstract and global, direct cross-attention between the two modalities may struggle to establish meaningful correspondences, making the alignment process inefficient.  
As a result, the cross-attention mechanism may focus on irrelevant regions or spread attention too broadly, leading to weak feature fusion and poor multimodal understanding. 

Conversely, applying deformable attention on the feature pyramid first enhances the multi-scale spatial structure of visual features before integrating textual information. 
Deformable attention selectively attends to the most relevant regions across different scales, improving efficiency and reducing redundancy in feature extraction \cite{zhu2020deformable}. 
By refining visual representations beforehand, the model provides a structured, semantically enriched foundation for cross-attention with text. 
This ensures the text interacts with meaningful, well-organized visual features rather than raw, noisy data. 
Consequently, the cross-modal alignment process becomes more effective, leading to better fusion of textual and visual information. 
Therefore, we propose CME, which can be formally defined as follows:
\begin{equation}
    I_n^\prime = \text{MSDeformAttn}(I_n, p_q, I_n)
\end{equation}
\begin{equation}
    E_n^{\prime l} = \text{Attn}(
    Q=\text{PE}^I(\textbf{I}_n^{\prime l}), 
    K=\text{PE}^S(\textbf{S}_w),
    V=\textbf{S}_w) \label{eq: F_fuse_cme}
\end{equation},
 and $E_n^\prime=\{E_n^{\prime 1}, \cdots, E_n^{\prime l}\}$.
 Here, $p_q$ are a set of reference points uniformly distributed on the feature pyramid.
 As depicted in \cref{fig: fuse}, each reference point \textcolor{orange}{\raisebox{-0.6ex}{\scalebox{2}{\textbullet}}} only attends to a small set of key sampling points \textcolor{purple}{\raisebox{-0.ex}{\scalebox{1.2}{$\blacktriangle$}}} around it across the multi-scale, regardless of the spatial size.
 The selection of key sampling points is learned via an MLP.

\subsection{Loss Function}
\label{subsec: loss}
The loss function can be decoupled as a spatial loss $\mathcal{L}^S$ and a spatial-temporal loss $\mathcal{L}^{Temp}$. 
$\mathcal{L}^S$ can be further decoupled as a \textit{track loss} $\mathcal{L}^T$ for existing targets and \textit{detect loss} $\mathcal{L}^D$ for newborn targets.
$\mathcal{L}^T$ is obtained via one-to-one computation between the tracking prediction triplet $\hat{y}_n^{Tra} = \{\hat{c}_n^{Tra}, \hat{b}_n^{Tra}, \hat{r}_n^{Tra}\}$ and the groundtruth $y_n^{Tra} = \{c_n^{Tra}, b_n^{Tra}, r_n^{Tra}\}$:
\begin{align}
    \mathcal{L}^T =& \lambda_{cls}\mathcal{L}_{cls}(\hat{c}_n^{Tra}, c_n^{Tra}) + \mathcal{L}_{box}(\hat{b}_n^{Tra}, b_n^{Tra}) + \notag \\ &\lambda_{ref}\mathcal{L}_{ref}(\hat{r}_n^{Tra}, r_n^{Tra})
\end{align}
, where $\mathcal{L}_{box}=\lambda_{L_1}\mathcal{L}_1+\lambda_{giou}\mathcal{L}_{giou}$ \cite{rezatofighi2019generalized}, $\mathcal{L}_{cls}$ and $\mathcal{L}_{ref}$ are focal loss \cite{lin2017focal}.
For $\mathcal{L}^D$, we find a bipartite graph matching, which of the predicted objects fits the true new-born objects.
Given the detection triplet $\hat{y}^{det}=\{\hat{c}_n^{Det}, \hat{b}_n^{Det}, \hat{r}_n^{Det}\}$ and the groundtruth $y_n^{Det} = \{c_n^{Det}, b_n^{Det}, r_n^{Det}\}$, we search for a permutation $\delta$ by minimizing matching cost:
\begin{align}
    \hat{\delta}=\arg\min\limits_{\delta}\mathcal{L}_{match}(\hat{y}_n^{Det}, y^{Det}_n(\delta))
\end{align},
where $\mathcal{L}_{match}=\mathcal{L}_{box}+\lambda_{cls}\mathcal{L}_{cls}$.
After determining $\hat{\delta}$, we use it as the new index of the predictions $y^{Det}_n(\hat{\delta})$ to compute $\mathcal{L}^{D}$:
\begin{align}
\mathcal{L}^D=\lambda_{cls}\mathcal{L}_{cls}+\mathbb{I}\lambda_{cls}\mathcal{L}_{cls}+\mathbb{I}\mathcal{L}_{box}
\end{align},
where $\mathbb{I}$ refers to $\mathbb{I}_{\{c^{Det}\neq \emptyset \}}$.
In addition, we include auxiliary decoding loss $\mathcal{L}_{aux}$ \cite{al2019character, carion2020end}, which calculates $\mathcal{L}^T$ and $\mathcal{L}^D$ using the intermediate outputs, after each decoder layer.
When adopting CQM, the $\hat \delta_{aux} = \arg\min\limits_\delta(\hat{y}_{n, aux_i}^{Det}, [y^{Tra}_n, y^{Det}_n](\delta))$ for the auxiliary losses, where $\hat{y}_{n, aux_i}^{Det}$ denotes the detection triplet of the $i$-th decoder layer (See \cref{fig: cola}).
$\mathcal{L}^{Temp}$, on the other hand, refines the bounding boxes using $\{\hat b_n^{Det\prime}, \hat b_n^{Tra\prime}\}$:
\begin{equation}
    \mathcal{L}^{Temp}=\mathcal{L}_{box}(\hat{b}_n^{Tra\prime}, b_n^{Tra})+\mathcal{L}_{box}(\hat{b}_n^{Det\prime}, b_n^{Det})
\end{equation}
The final loss is:
\vspace{-1em}
\begin{align}
    \mathcal{L} = \mathcal{L}^D + \mathcal{L}^T + \mathcal{L}^{Temp} +\sum\limits_{i=0}^{N_{aux}}\mathcal{L}_{aux_i}
\end{align},
\vspace{-1em}
where $N_{aux}$ is the number of decoder layers.

\section{Experiments} \label{sec: exp}
\begin{figure}[t]
  \centering
  \includegraphics[width=1\linewidth]{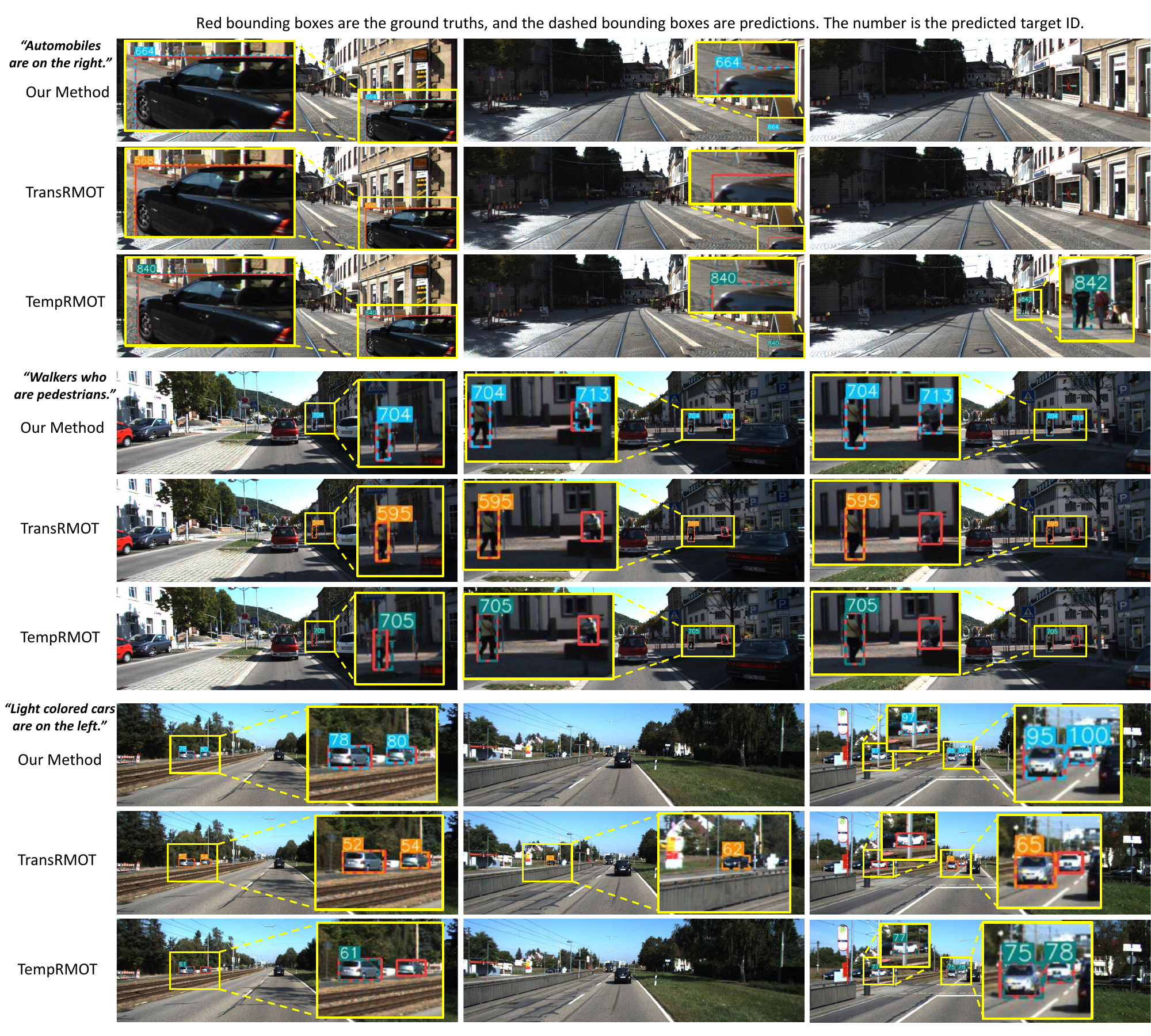}
  \vspace{-2em}
   \caption{Qualitative comparison. Red bounding boxes are the ground truths, and the dashed bounding boxes are predictions. Compared with the baselines, our method (\Mname) achieves more accurate localization and ID association, with fewer false positives and negatives. Zoom in on the figure for more details.}
   \label{fig: qualitative comparison}
   \vspace{-2em}
\end{figure}

\subsection{Experimental Setup}
\textbf{Datasets.} We evaluate the proposed method on two datasets: Refer-KITTI \cite{wu2023referring} and Refer-KITTI-V2 \cite{zhang2024bootstrapping}.

\noindent \textbf{Evaluation Metrics.}  To ensure fair comparison with prior baseline~\cite{wu2023referring}, we also employ Higher Order Tracking Accuracy (HOTA) \cite{luiten2021hota} as the primary evaluation metric.
HOTA measures the alignment between the predicted and ground-truth trajectories. 
It provides a comprehensive and balanced assessment by jointly considering the performance of detection and association. It is defined as the geometric mean of detection accuracy (DetA) and association accuracy (AssA), i.e., $\text{HOTA}=\sqrt{\text{DetA}\cdot\text{AssA}}$.
Additionally, we adopt the following sub-metrics: detection recall/precision (DetRe/DetPr), association recall/precision (AssRe/AssPr), and
localization accuracy score (LocA).

\begin{table*}[t]
\centering
\vspace{-.5em}
\resizebox{1\textwidth}{!}{%
\begin{threeparttable}
\begin{tabular}{c|c|cccccccc|cccccccc}
\toprule
Method &
  E\tnote{a} &
  \multicolumn{8}{c|}{\textbf{Refer-KITTI}} &
  \multicolumn{8}{c}{\textbf{Refer-KITTI-V2}} \\ \midrule
 &
   &
  \textbf{HOTA} $\uparrow$ &
  DetA$\uparrow$ &
  AssA$\uparrow$ &
  DetRe$\uparrow$ &
  DetPr$\uparrow$ &
  AssRe$\uparrow$ &
  AssPr$\uparrow$ &
  LocA$\uparrow$ &
  \textbf{HOTA}$\uparrow$ &
  DetA$\uparrow$ &
  AssA$\uparrow$ &
  DetRe$\uparrow$ &
  DetPr$\uparrow$ &
  AssRe$\uparrow$ &
  AssPr$\uparrow$ &
  LocA$\uparrow$ \\ \midrule
FairMOT \cite{zhang2021fairmot} \footnotesize{IJCV21} &
  \ding{55} &
  22.78 &
  14.43 &
  39.11 &
  16.44 &
  45.48 &
  43.05 &
  71.65 &
  74.77 &
  22.53 &
  15.80 &
  32.82 &
  20.60 &
  37.03 &
  36.21 &
  71.94 &
  78.28 \\
  
ByteTrack \cite{zhang2022bytetrack}  \footnotesize{ECCV22}&
  \ding{55} &
  24.95 &
  15.50 &
  43.11 &
  18.25 &
  43.48 &
  48.64 &
  70.72 &
  73.90 &
  24.59 &
  16.78 &
  36.63 &
  22.60 &
  36.18 &
  41.00 &
  69.63 &
  78.00 \\
iKUN \cite{du2024ikun} \footnotesize{CVPR24}&
  \ding{55} &
  48.84 &
  35.74 &
  66.80 &
  51.97 &
  52.25 &
  72.95 &
  87.09 &
  - \tnote{b} &
  10.32 &
  2.17 &
  49.77 &
  2.36 &
  19.75 &
  58.48 &
  68.64 &
  - \tnote{b} \\
  
  ReferGPT \cite{chamiti2025refergpt} \footnotesize{CVPR25}&
  \ding{55} &
  46.36 &
  36.58 &
  59.00 &
  51.40 &
  52.16 &
  73.16 &
  36.31 &
  83.26 &
  30.12 &
  15.69 &
  {\color[HTML]{FE0000} 59.02} &
  21.55 &
  34.41 &
  {\color[HTML]{FE0000} 74.59} &
  68.20 &
  79.76 \\
  
TransRMOT \cite{wu2023referring} \footnotesize{CVPR23}&
  \ding{51} &
  46.56 &
  37.97 &
  57.33 &
  49.69 &
  60.10 &
  60.02 &
  89.67 &
  90.33 &
  31.00 &
  19.40 &
  49.68 &
  36.41 &
  28.97 &
  54.59 &
  82.29 &
  89.82 \\
TempRMOT \cite{zhang2024bootstrapping} &
  \ding{51} &
  52.21 &
  40.95 &
  66.75 &
  55.65 &
  59.25 &
  71.82 &
  87.76 &
  90.40 &
  35.04 &
  22.97 &
  53.58 &
  34.23 &
  40.41 &
  59.50 &
  81.29 &
  90.07 \\

HFF-Tracker \cite{zhao2025hff}  \footnotesize{AAAI25}&
  \ding{51} &
  52.41 &
  41.29 &
  66.65 &
  53.42 &
  {\color[HTML]{FE0000} 62.89} &
  71.48 &
  88.96 &
  {\color[HTML]{FE0000} 90.76} &
  36.18 &
  {\color[HTML]{FE0000} 24.64} &
  53.27 &
  36.86 &
  41.83 &
  59.42 &
  81.40 &
  89.77 \\

CDRMT \cite{liang2025cognitive} \footnotesize{Information Fusion}&
  \ding{51} &
  49.35 &
  40.34 &
  60.56 &
  54.54 &
  59.30 &
  64.70 &
  89.80 &
  90.61 &
  31.99 &
  20.37 &
  50.35 &
  26.40 &
  {\color[HTML]{FE0000}46.26} &
  53.40 &
  85.90 &
  90.36 \\

Ours (\mname) &
   \ding{51} &
   {\color[HTML]{FE0000} \textbf{55.63}} &
   {\color[HTML]{FE0000} \textbf{43.73}} &
   {\color[HTML]{FE0000} \textbf{70.77}} &
   {\color[HTML]{FE0000} \textbf{62.68}} &
   59.12 &
   {\color[HTML]{FE0000} \textbf{74.52}} &
   {\color[HTML]{FE0000} \textbf{93.39}} &
   89.83 & {\color[HTML]{FE0000} \textbf{37.67}}
   & 24.09
   & 58.92
   & {\color[HTML]{FE0000} \textbf{42.87}}
   & 35.95
   & 63.91
   & {\color[HTML]{FE0000} \textbf{87.48}}
   & {\color[HTML]{FE0000} \textbf{91.59}}
   \\ \bottomrule
\end{tabular}%
\begin{tablenotes}[leftmargin=0.35cm, itemindent=.0cm, itemsep=0.0cm, topsep=0.1cm]
\item[a] 
``E'' means end-to-end training.
\item[b] iKUN conducts oracle experiments, i.e., the bounding boxes are revised based on ground truth. 
\end{tablenotes}
\end{threeparttable}
}
\vspace{-1em}
\caption{Quantitative results of our method and the baselines. The best performance is highlighted in red.}
\label{tab: hota}
   \vspace{-1.7em}
\end{table*}

\noindent \textbf{Model Details.} We leverage the same backbone and text encoder as \cite{wu2023referring} to extract both image embeddings and linguistic embeddings. 
As with deformable DETR \cite{zhu2020deformable}, we adopt the last four feature maps of the backbone as the input to the CME.
The parameters associated with the CME are initialized with random values, and the parameters of the text encoder are frozen during training.
The remaining parameters are initialized with official pre-trained weights from \cite{zhu2020deformable} on the COCO dataset \cite{lin2014microsoft}.
Our optimization employs AdamW with a base learning rate of $10^{-4}$, except for the visual backbone with a learning rate of $10^{-5}$. 
Beginning from the $40^{th}$ epoch, we decrease the learning rate by a factor of 10. 
The window length $K$ for temporal reasoning is set to 5.
We conduct end-to-end training on 6 NVIDIA RTX A6000 GPUs.
During inference, the model operates without the need for post-processing, such as non-maximum suppression \cite{canny1986computational}. 
We employ detection thresholds $\beta_{obj} = 0.7$ and a referring threshold $\beta_{ref} = 0.3$ to localize visible objects and filter referent targets.
\subsection{Qualitative Results} \label{subsec: qualitative}
We visualize some examples in \cref{fig: qualitative comparison}. 
TellTrack consistently demonstrates superior performance, exhibiting robust tracking and high localization accuracy across the cases. 
Its predictions maintain stable associations with the correct targets, such as the ``automobiles on the right'' and the ``light colored cars on the left''. In contrast, the baselines exhibit notable failure modes. The baselines are prone to generating false positives, incorrectly identifying non-target objects and background elements, and struggle when faced with objects that are small and densely distributed (``Walkers who are pedestrians'').
This visual evidence suggests that TellTrack outperforms the baselines in both target localization and long-term association stability.

\subsection{Quantitative Results}\label{subsec: quantitative}

We examine the proposed method and several competitors in \cref{tab: hota}. 
For the \textit{``detect-and-track''} methods, i.e., FairMOT\cite{zhang2021fairmot}, ByteTrack\cite{zhang2022bytetrack}, we integrated the encoder into the detection module, followed by independent trackers to associate each referent target, for a fair comparison. 
iKUN \cite{du2024ikun} adopts the same paradigm exploiting a foundation model CLIP \cite{radford2021learning} to adaptively extract visual features.
For the \textit{one-stage} methods, we compare \mname with TransRMOT \cite{wu2023referring}, TempRMOT \cite{zhang2024bootstrapping}, HFF-Tracker \cite{zhao2025hff}, and CDRMT \cite{liang2025cognitive}.
On both datasets, our method achieves a superior performance (HOTA of 55.63\% on Refer-KITTI, and 37.67\% on Refer-KITTI-V2).
Specifically, we surpass the previous best model by a significant margin of 3.22\% and 1.49\% on two datasets.
\subsection{Ablation Study}\label{subsec: ablation}
To investigate the effect of core components in \Mname, we conduct extensive ablation studies on Refer-KITTI-V2.
\Cref{tab:ablation} illustrates the results of all combinations of our proposed components. 
Every combination exhibits a positive impact on the overall performance.
Specifically, using CME effectively fuses the information from different modalities, remarkably improving the association (+2.84\% on AssA).
CQM, on the other hand, improves detection and association simultaneously, thanks to more activations for the detection queries during training.
RIQA provides explicit intention guidance to the queries, boosting the reasoning ability of the model and, therefore, leading to a significant improvement in association (+3.32\% on AssA, and +8.02\% on AssPr).
We further investigate two types of RIQA, as shown in \cref{tab:ablation-riqa}.
For pre-decoder adaptation, we examine the effect of infusing sentence embedding with detection queries, track queries, and both.
The results indicate that both pre-decoder and in-decoder RIQA improve performance compared to decoding by vanilla queries.

\begin{figure}[t]
  \centering
  \includegraphics[width=1\linewidth]{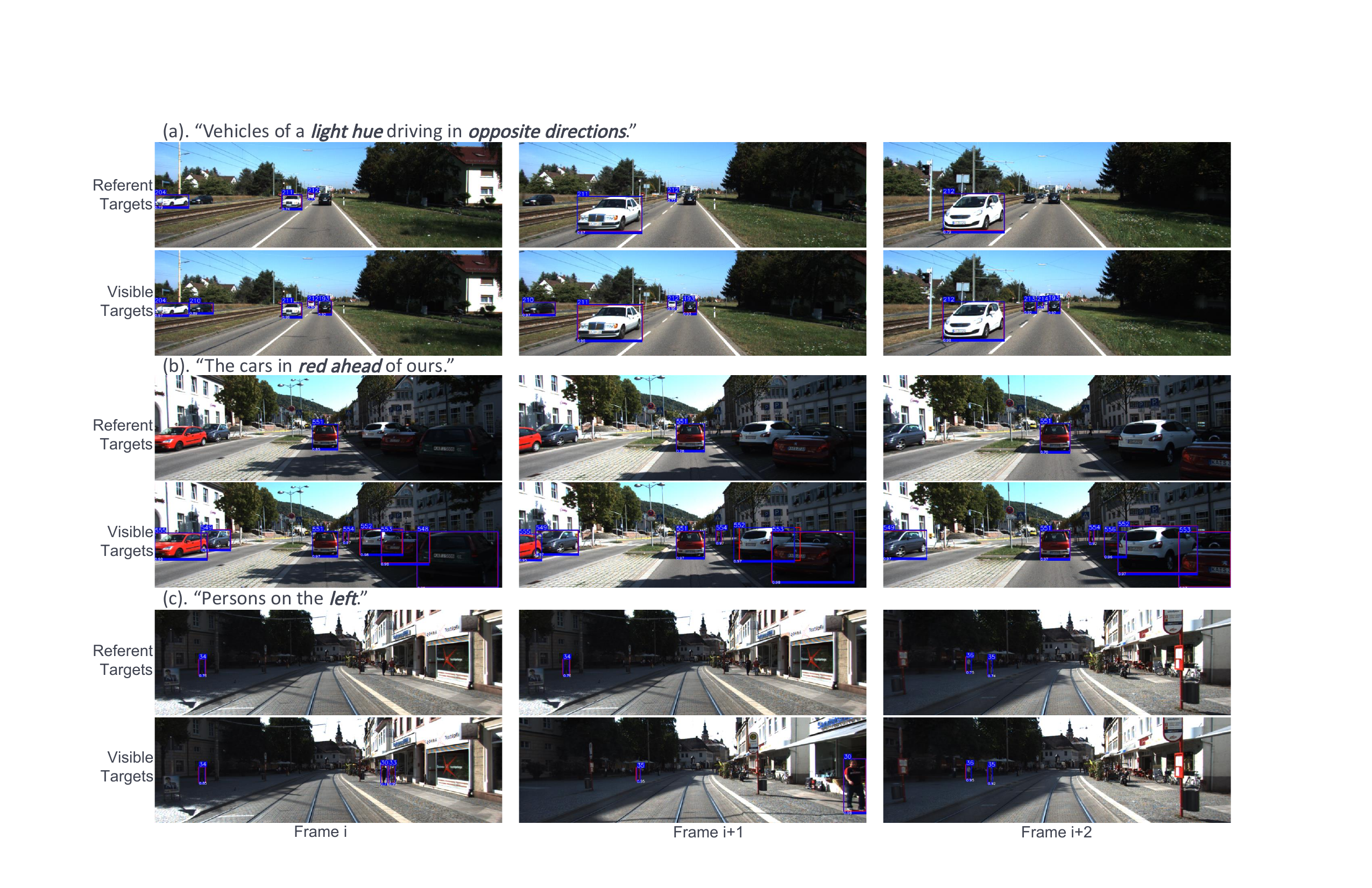}
   \caption{The bottom panels show all visible objects detected by \Mname. Blue bounding boxes stand for model predictions, and the red bounding boxes are the ground truths. The number above each prediction is the assigned ID for that target, and the same ID across each frame represents the active tracking. The number below each prediction is the referring score/ object score, respectively. Zoom in on the figure for more details.}
   \label{fig: qualitative}
\end{figure}

In addition, we investigate the effect of $\beta_{ref}$ in \cref{tab:ablation-ref-thres} and $\beta_{obj}$ in \cref{tab:ablation-det-thres}. Depending on the thresholds, the model predicts whether an object is detected and fits the referring text.
Overall, both thresholds have a significant impact on the balance between precision and recall. 
\mname achieves the best performance when $\beta_{ref}=0.3, \beta_{obj}=0.7$.
We visualize some examples to show the object detection and referring association performance of \mname in~\cref{fig: qualitative}.
The upper panels visualize the predicted tracked referent targets by \Mname, and the lower panels show all detected visible objects by \Mname.
As depicted, \mname precisely understands and recognizes the meaning of object category, color, and position intentions in the text prompts, identifies and tracks the referent targets accurately, even in various challenging situations, such as multiple objects, change of object status, and varying number of instances.
In \cref{fig: qualitative} (a), \mname successfully identifies the concepts ``vehicles with light color'', and ``opposite direction''. 
In \cref{fig: qualitative} (b), the model understands the meaning of ``red'' and ``ahead of us'', tracking the red car with ID 551, while filtering out the red car with ID 555 on the left.
Similarly, the model successfully only tracks the people on the left while filtering out all people on the right in~\cref{fig: qualitative} (c).

Lastly, we examine the inference speed of \Mname, crucial for real-world applications.
While the baseline, TempRMOT, achieves 15.13 FPS, \mname maintains a comparable 15.08 FPS with improved performance.

\begin{table}[t]
\resizebox{\columnwidth}{!}{%
\begin{tabular}{ccc|cccccccc}
\toprule
\multicolumn{3}{c|}{Components} & \multicolumn{8}{c}{Metrics}                               \\ \midrule
RIQA       & CQM       &    CME    & HOTA$\uparrow$ & DetA$\uparrow$ & AssA$\uparrow$ & DetRe$\uparrow$ & DetPr$\uparrow$ & AssRe$\uparrow$ & AssPr$\uparrow$ & LocA$\uparrow$ \\ \midrule
           &           &           & 35.04& 22.97& 53.58&  34.23&  {\color[HTML]{FE0000}40.41}&  59.50&  81.29&  90.07\\
           &           & \ding{51} & 35.66& 22.54& 56.42&  {\color[HTML]{FE0000}45.99}&  30.65&  63.62&  81.77&  90.98\\
           & \ding{51} &           & 36.06& 23.05& 56.43&  39.20&  35.87&  61.80&  85.92&  91.08\\
           & \ding{51} & \ding{51} & 36.27& 23.16& 56.80&  41.02&  34.73&  63.52&  81.67&  91.58\\
\ding{51}  &           &           & 36.53& 23.83& 56.02&  38.89&  38.07&  61.86&  87.41&  90.06\\
\ding{51}  &           & \ding{51} & 36.45& 23.40& 56.71&  38.61&  37.35&  61.71&  87.22&  90.76\\
\ding{51}  & \ding{51} &           & 36.77& 23.96& 56.41&  42.38&  35.54&  61.83&  86.16&  90.20\\
\ding{51}  & \ding{51} & \ding{51} & {\color[HTML]{FE0000}37.67}& {\color[HTML]{FE0000}24.09}& {\color[HTML]{FE0000}58.92}&  42.87&  35.95&  {\color[HTML]{FE0000}63.91}&  {\color[HTML]{FE0000}87.48}&  {\color[HTML]{FE0000}91.59}\\ \bottomrule
\end{tabular}%
}
\vspace{-.5em}
\caption{Ablation studies of different components in \Mname.}
\vspace{-1em}
\label{tab:ablation}
\end{table}

\begin{table}[t]
\resizebox{\columnwidth}{!}{%
\begin{tabular}{cc|cccccccc}
\toprule
\multicolumn{2}{c|}{Method}        & \multicolumn{8}{c}{Metrics} \\ \midrule
\multicolumn{2}{c|}{1. Pre-decoder} &
  \multirow{2}{*}{HOTA$\uparrow$} &
  \multirow{2}{*}{DetA$\uparrow$} &
  \multirow{2}{*}{AssA$\uparrow$} &
  \multirow{2}{*}{DetRe$\uparrow$} &
  \multirow{2}{*}{DetPr$\uparrow$} &
  \multirow{2}{*}{AssRe$\uparrow$} &
  \multirow{2}{*}{AssPr$\uparrow$} &
  \multirow{2}{*}{LocA$\uparrow$} \\
$Q^{Detect}$     & $Q^{Track}$     &   &   &   &   &   &   &  &  \\ \midrule
                 &  \ding{51}      &35.45&22.77&55.19&50.08&29.46&{\color[HTML]{FE0000}62.65}&80.80&85.14\\
  \ding{51}      &                 &35.88&{\color[HTML]{FE0000}23.73}&54.26&{\color[HTML]{FE0000}51.78}&30.46&61.59&80.32&82.58\\
  \ding{51}      &  \ding{51}      &{\color[HTML]{FE0000}36.23}&23.07&{\color[HTML]{FE0000}56.90}&34.99&{\color[HTML]{FE0000}40.37}&60.97&{\color[HTML]{FE0000}89.31}&{\color[HTML]{FE0000}90.55}\\ \midrule
\multicolumn{2}{c|}{2. In-decoder} &35.37&22.76&54.95&37.48&36.69&60.22&85.70&90.16\\ \bottomrule
\end{tabular}%
}
\caption{Effects of different RIQA. For pre-decoder adaptation, the sentence embedding can be infused with either detection queries, track queries, or both.}
\label{tab:ablation-riqa}
\end{table}

\begin{table}[t]
\resizebox{\columnwidth}{!}{%
\begin{tabular}{c|cccccccc}
\toprule
$\beta_{ref}$ & HOTA & DetA & AssA & DetRe & DetPr & AssRe & AssPr & LocA \\ \midrule
0.2   & 37.06& 23.80& 57.72& {\color[HTML]{FE0000}43.06} & 34.73 & 64.30 & 82.74 & 91.03   \\
0.3   & {\color[HTML]{FE0000}37.22}& 23.38& {\color[HTML]{FE0000}59.25}& 39.12 & 36.74 & {\color[HTML]{FE0000}64.85} & 86.92 & 91.08\\
0.4   & 36.84& {\color[HTML]{FE0000}23.86}& 56.88& 39.71 & 37.42 & 62.69 & 83.66 & 91.10     \\
0.5   & 36.03& 23.22& 55.91& 36.49 & 38.97 & 60.97 & 84.30 & 91.17     \\
0.6   & 34.94& 22.07& 55.33& 32.56 & 40.65 & 59.62 & 85.46 & 91.26\\
0.7   & 33.39& 20.45& 54.54& 27.93 & 43.27 & 58.11 & 86.91 & 91.47     \\
0.8   & 29.99& 17.15& 52.45& 21.17 & {\color[HTML]{FE0000}47.47} & 55.09 & {\color[HTML]{FE0000}88.72} & {\color[HTML]{FE0000}91.85}\\ \bottomrule
\end{tabular}
}
\caption{Performance of \mname on different $\beta_{ref}$.}
\label{tab:ablation-ref-thres}
\end{table}

\begin{table}[h]
\centering
\resizebox{\columnwidth}{!}{%
\begin{tabular}{c|cccccccc}
\toprule
$\beta_{obj}$ & HOTA & DetA & AssA & DetRe & DetPr & AssRe & AssPr & LocA \\ \midrule
0.3   &34.47&22.36&53.32&{\color[HTML]{FE0000}39.79}&33.36&59.34&81.84&90.36\\
0.7   & {\color[HTML]{FE0000}37.22}& {\color[HTML]{FE0000}23.38}& {\color[HTML]{FE0000}59.25}& 39.12 & 36.74 & {\color[HTML]{FE0000}64.85} & 86.92 & {\color[HTML]{FE0000}91.08}     \\
0.9   & 23.14& 11.41& 46.99& 12.98 & {\color[HTML]{FE0000}47.95} & 48.86 & {\color[HTML]{FE0000}91.05}& 90.09\\ \bottomrule
\end{tabular}
}
\caption{Performance of \mname on different $\beta_{obj}$.}
\label{tab:ablation-det-thres}
\vspace{-1em}
\end{table}

\section{Related Work}
\vspace{-.5em}
\subsection{Referring Understanding}
The core challenge of referring understanding is to model the semantic alignment of cross-modal sources.
Early methods \cite{mao2016generation, liu2019learning, chen2024taskclip} mainly fuse the sources in two stages: 1). Adopting an off-the-shelf object detector to propose massive object proposals. 2). Leveraging a semantic alignment model to learn the similarity between proposals and language expressions and find the best-fitted objects.
Nevertheless, the performance of these methods heavily relies on the quality of the object detector.
While recent referring expression comprehension (REC) models (e.g., VISA~\cite{yan2024visa}, HyperSeg~\cite{wei2024hyperseg}, GLUS~\cite{lin2025glus}) address video REC, they are limited to single-object grounding, require multiple video passes, and operate offline. 
In contrast, we tackle RMOT under streaming (causal) conditions, handling challenges such as detecting multiple newborn targets, disappearance/reappearance, and online association, which are beyond the scope of REC.

\subsection{Multi-Object Tracking}
Prior works \cite{braso2020learning, bergmann2019tracking, dendorfer2021motchallenge} adopt a two-stage \textit{detect-and-track} paradigm.
They first detect objects in each frame, and then associate the detections across frames, thereby tracking individual objects over time.
Recent works \cite{zhang2023motrv2, meinhardt2022trackformer, zeng2022motr} propose \textit{one-stage} trackers, mostly based on trainable transformer \cite{vaswani2017attention} encoder-decoder architecture.
They formulate MOT as a set prediction problem, by representing objects implicitly in the decoder queries, which are embeddings used by the decoder to output bounding box coordinates and class predictions.
To further improve the performance, some efforts investigate the use of temporal memory \cite{cai2022memot}, domain adaptation \cite{yu2023generalizing}, and label reassignment \cite{yu2023motrv3, yan2023bridging}.
\vspace{-.5em}

\subsection{Referring Tracking}
\vspace{-.5em}
Referring SOT has been
studied for several years. 
Most recent SOTA solutions mainly follow the
joint tracking paradigm.
MTTR \cite{botach2022end} applies a DETR-like \cite{carion2020end} multi-modal
module to decode instance-level features into a set of multimodal sequences. 
ReferFormer \cite{wu2022language} inputs a set of object queries conditioned on language descriptions into a Transformer to estimate the referred object.
As for RMOT, along with Refer-KITTI, the first baseline model, TransRMOT \cite{wu2023referring} is introduced.
It is built upon the end-to-end multi-object
tracking method MOTR \cite{zeng2022motr} to accept the cross-modal input. 
The latter approach, iKUN \cite{du2024ikun}, follows a two-stage paradigm. 
It first explicitly extracts object proposals and then selects the objects matched with the language expression.
On the other hand, it introduces a neural version of the Kalman filter to dynamically adjust process noise and observation noise based on the current motion status.
On top of TransRMOT, TempRMOT \cite{zhang2024bootstrapping} integrates historical information into the model, refining the predictions with the help of the outputs from previous frames.

\vspace{-1em}
\section{Conclusion}
\vspace{-.5em}
We present a novel end-to-end framework RMOT algorithm, \Mname.
We introduce a new matching strategy during training, which effectively alleviates the imbalanced activations between detection queries and track queries caused by the difference in the number of newborn targets and existing targets in the dataset.
We also propose a query adaptation component that explicitly fuses the linguistic intention with the decoder queries, and enhances the reasoning.
Besides, we redesign the encoder to improve multi-modal fusion.
\mname is evaluated on the widely used datasets and achieves the SOTA performance, demonstrating the effectiveness of our proposed components.
{
    \small
    \bibliographystyle{ieeenat_fullname}
    \bibliography{main}
}

\end{document}